# Real-time Human-Robot Collaborative Manipulations of Cylindrical and Cubic Objects via Geometric Primitives and Depth Information

Huixu Dong[1], Jiadong Zhou[2], Haoyong Yu[1]

*Abstract*—Many objects commonly found in household and industrial environments are represented by cylindrical and cubic shapes. Thus, it is available for robots to manipulate them through the real-time detection of elliptic and rectangle shape primitives formed by the circular and rectangle tops of these objects. We devise a robust grasping system that enables a robot to manipulate cylindrical and cubic objects in collaboration scenarios by the proposed perception strategy including the detection of elliptic and rectangle shape primitives and depth information. The proposed method of detecting ellipses and rectangles incorporates a one-stage detection backbone and then, accommodates the proposed adaptive multi-branch multi-scale net with a designed iterative feature pyramid network, local inception net, and multi-receptive-filed feature fusion net to generate object detection recommendations. In terms of manipulating objects with different shapes, we propose the grasp synthetic to align the grasp pose of the gripper with an object's pose based on the proposed detector and registered depth information. The proposed robotic perception algorithm has been integrated on a robot to demonstrate the ability to carry out human-robot collaborative manipulations of cylindrical and cubic objects in real-time. We show that the robotic manipulator, empowered by the proposed detector, performs well in practical manipulation scenarios. (An experiment video is available in YouTube, https://www.youtube.com/watch?v=Amcs8lwvNK8 .)

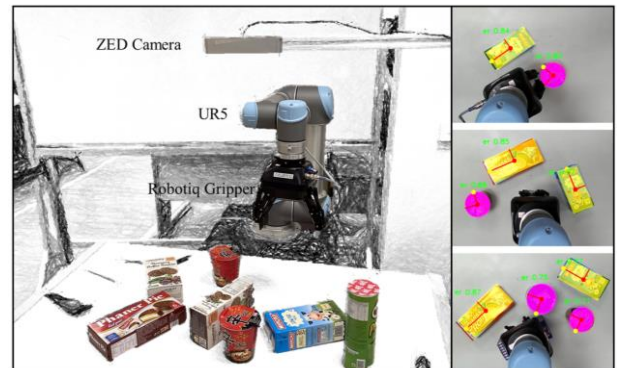

Figure. 1. The robot grasping cylindrical and cubic objects. The pink ellipses represent the cylindrical objects and yellow rectangles indicate the cubic objects; the green text is the object class confidence.

## I. INTRODUCTION

Robots manipulating objects and transfer of objects between humans and robots are critical capabilities for collaborative robots across a multitude of application areas from physical assistance in household environments to collaborative manufacturing in industrial scenarios[1-3]. Examples of robot-human collaborative manipulations include a household robot that assists its user to grasp daily objects, a collaborative industrial robot that passes parts to a human worker for assembly, and so on. One difficulty of such physical activities is to enable robots to perform successful grasps in real-time, which is the main feature of human-robot collaboration manipulation.

A great number of objects in households and industries, such as cans, cups, cellphones, toothpaste boxes, circle or rectangle mechanical parts, have surfaces with elliptic and rectangle geometric primitives so that cylindrical and cubic objects in images are represented through ellipses and rectangles depicting the circular tops of the cylindrical objects at most angles of observation and the rectangle tops of cubic objects under the view of vertical angles, respectively. Therefore, a robot can grasp cylindrical and cubic objects via detecting ellipse and rectangle geometric primitives. In this work, we focus on real-time human-robot collaborative manipulations of cylindrical and cubic objects by the proposed detector of ellipses and rectangles and depth information in practical scenarios; see Fig.1 for an example.

Indeed, ellipse detection has been deemed useful in the computer vision applications, such as industrial inspection, medical diagnosis, security, autonomous navigation, and tracking targets[4]. Moreover, rectangle detection arises in several practical applications, such as text box detection, tracking vehicles, and recognition of buildings[5]. The prior detection algorithms of ellipses or rectangles always cause failure detections, especially in cluttered scenes where target objects are often densely placed. In different observation views, a variety of scales for objects results in the shape variations of ellipses and rectangles. Moreover, the low detection speed of an ellipse or a rectangle poses a considerable challenge for ellipse and rectangle detection. Almost all the detection approaches of ellipse or rectangle are not available for robotic static and dynamic manipulation because of low shape detection accuracy or execution speed, sometimes both. We note here that Dong et al.[1] successfully realized robotic static and dynamic manipulation of cylindrical objects via the proposed real-time accurate ellipse detection and also, have a robust capability of detecting small, occluded, and overlapped ellipses. However, this method has an unsatisfactory performance of detecting small ellipses in practical scenarios. In addition, this cannot detect rectangles.

To address or relieve the above issues for achieving human-robot collaborative manipulations of cylindrical and cubic objects, we propose a novel strategy via a constructed end-to-end network framework of fast and accurate detection of ellipses and rectangles (Note that here ellipses includes circles and rectangles includes squares.) in practical scenarios. In particular, we first extract original features from the input image with basic geometric primitives through the one-stage backbone[6], then accommodate the proposed adaptive multibranch multiscale net with a designed iterative feature pyramid network, local inception net, and multi-receptive-filed feature fusion net to generate object detection recommendations and determine object position. We design a new loss function, considering the position of the rotated bounding box and object class. Finally, we utilize a highly fast modified rotation R-NMS method to determine the final detection.

[1]Bio-Robotics Lab, National University of Singapore, Singapore 117583, Singapore; [2]Robotics Research Center, Nanyang Technological University, Singapore 639798, Singapore. Email: bieyhy@nus.edu.sg.

We **highlight** the **novelties** of this work. Foremost, our core contribution of this work is addressing the problem of perceiving cylindrical and cubic objects for real-time robotic manipulations via elliptic and rectangle geometric primitives in practical scenarios. The work provides the first solution, to the best of our knowledge, for robotic manipulation of cylindrical and cubic objects in physical scenarios through elliptic and rectangle shape primitives only. The second novelty locates in the detection of ellipses and rectangles. We are the first to propose the incorporation of detecting four shapes with different geometric primitives-circle, ellipse, square, rectangle and achieve segmented-fitting detection simultaneously rather than just use a bounding box to indicate a zone covering the detected objects, which provides many probabilities of robotic practical applications. Also, the proposed model takes the combination advantage of the same data structure representatives of ellipse and rectangle in the training data to simplify the model design such as to detect ellipses and rectangles. The third novelty incorporates several new aspects. First, an end-to-end one-stage detection model for multi-scale ellipses and rectangles is constructed. The newly proposed model archives an effective detection performance on ellipses and rectangles at a sufficiently fast speed in cluttered environments. An adaptive multi-branch multi-scale module(AMM) is proposed to fuse multilevel feature reuse for enhancing feature propagation. Second, a wide variety of rotated anchor boxes and a new loss function are designed to match targets of a large number of scales of ellipses and rectangles. There is a small contribution that we propose an approach of building synthetic datasets via compositing the real background images and 3D models of grasped objects and automatically generating transferred label points to bridge the domain gap of the insufficient quantity of the manually labeled data, thereby increasing the robustness of the network. The last contribution is that we implement our perception strategy based on the ellipse and rectangle detection on an industrial robot for achieving plenty of human-robot collaborative manipulations in cluttered scenarios, obtaining highly successful rates of grasp experiments.

The work is the **extension** of the robotic grasp strategy proposed in [1]. Although the work [1] also found the importance of considering the real-time ellipse detector as a perception strategy for robotic manipulations in complex scenarios, the underlying motivation, methodology as well as purpose in this paper are quite different. The rest of this paper is organized as follows. The following section II introduces the related work. Section III presents the proposed algorithm for the detection of ellipse and rectangle. Section IV provides demonstrations of robot-human collaborative manipulation scenarios. The paper is concluded in Section V.

## II. RELATED WORK

A vast amount of prior works on the topics of geometric shape and robotic grasp detections were developed. Here we just introduce some representative works with respect to our work.

### A. Geometric Shape Detection
*1) Ellipse detection*

Until now, almost all ellipse detection methods are based on parametric analysis, algebraic analysis, as well as geometric analysis of the properties of ellipses[1, 7]. Parametric analysis approaches [8, 9] based on Hough Transform were developed to detect ellipses through combing with geometric or algebraic analysis[7, 10, 11]. The algebraic approaches, such as those utilizing least-squares fitting approaches [12, 13], random sample consensus [14], etc., extract edges that are likely to consist of detected ellipses. Moreover, there exist some geometry-based ellipse detection methods[1, 7, 11] that utilize short straight lines to approximate arc segments and then, these arc segments are grouped and merged into ellipses. Nevertheless, they are still limiting for rapidly detecting ellipses in severely complex environments.

*2) Rectangle detection*

For rectangle detection, some of the traditional analytic techniques explore geometric characteristics of a rectangle in Hough Transform space[15], ignoring the extraction and/or grouping of linear segments [5]. However, most traditional rectangle detection methods illustrated in the literature capture the information from edge[16], line[17, 18], and corner primitives[19, 20]. There are a lot of learning-based approaches regressing the bounding box of the detection box. However, the shapes of many objects are not rectangle strictly and the predicted bounding box just can cover the profile of the object. It does not indicate that the covered object is with rectangle geometric primitives. That is, regressing a rectangle is just considered as the detection of a bounding box. CNN-based algorithms of the regression bounding box are applied to detecting the ships in remote sensing images and text boxes [21-24]. Some methods were proposed to detect the horizontal box bound of a ship in remote sensing images[25]. Compared with the ship detection methods, works on text detection [23, 24] generally pay more attention to object orientation. However, these approaches are generally invalid in handling images in complex environments where ships and boats in remote sensing images and scene texts are small, occluded by each other, or some detectors cannot reach real-time requirements.

### B. Robotic Grasp Detection of Objects

Robotic grasp detection of objects is a widely studied field in robot applications. These detection methods include traditional analytical methods and learning-based approaches[26]. Early traditional grasp detection makes use of a set of 2-dimensional(2D) key points and the edge features to obtain the information of the class and position of the targeted object [27, 28]. For the robotic grasp detection, the 3-dimensional(3D) descriptor captured by a RGB-D sensor is available and more feasible than ever. These works based on the 3D descriptor further combines local template descriptor, shape, depth, and scene semantic as features to detect the objects[29]. Recently, learning-based methods (e.g.,[30] [31],[32], [33], [26]) have proven to produce an effective performance for robot grasping detection, mapping from the features-set to grasp-quality via extracting rich features from the training data. In this context, most of these methods regressed oriented rectangles (defined by their positions, orientations, widths, and heights) in scenarios to do grasp detection representatives. The above detection methods have difficulty in coping with these more challenging robotic application scenarios where a robot utilizes the above detection methods to rapidly manipulate objects in dynamic scenarios. In addition, such grasp detection ignores the object itself shape and does not fit the profile of the grasped object, which results in invalid robotic manipulation such as assembling tasks.

### III. THE PROPOSED METHODOLOGY

The approach for the detection of elliptic and rectangle shapes in images consists of three main blocks, namely, data preparation, network architecture construction, and detail implementation.

#### A. Data Preparation

We establish a synthetic dataset with cylindrical and cubic objects for robotic grasp(S-RG) for evaluating the proposed method for robotic grasp. The key reason is that generating synthetic datasets can relieve human labor. We elaborately choose several daily objects as a representative set, including the cylindrical and cubic objects, and employ a 3D scanner-Artec Eva to scan 3D models of objects, as shown in Fig.2(A). A virtual camera in OpenGL is used in rendering and capturing the target object images from various viewpoints uniformly (see Fig.2-B). Sequentially, the scene images are collected as the backgrounds. Then, the background image and the object images are composited to generate images of the virtual scenes with the target objects expected to be encountered in the real physical world. In terms of S-RG dataset, we capture images from practical grasping scenarios as backgrounds.

#### B. Learning Model Construction
##### 1) Network Architecture

The end-to-end framework of the proposed method is summarized in Fig.3, which mainly consists of two parts. Specifically, the proposed network first employs the YOLOv3-Darknet53 architecture [6] as a backbone network to extract original features from the input image for the detection of objects with basic geometric primitives, then accommodate the proposed adaptive multibranch multiscale net (AMM-Net) with designed iterative feature pyramid fusion network (IFPF), local inception net and multi-receptive-filed feature fusion net to generate object detection recommendations and determine object position by predicting

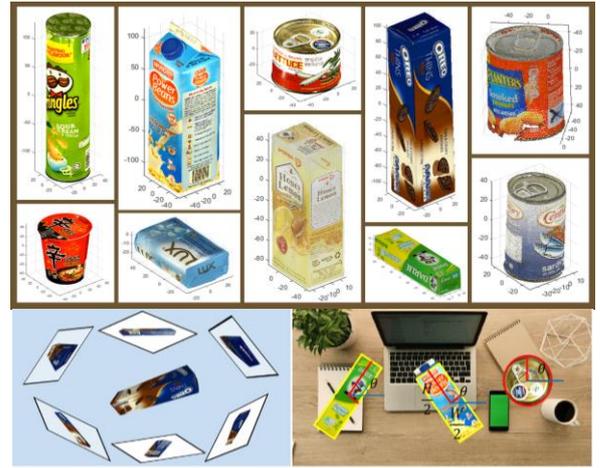

Figure.2. The 3D models of some cylindrical and cubic objects (A). A virtual camera captures images from the 3D models and the images extracted from 3D models; the synthetic images extracted from 3D models of objects and practical background; the cylindrical and cubic objects are described by five parameters (B). $H, W$ and $\theta$ represent the height, width and orientation of the object, respectively.

hull orientation and relative offset.

##### a) Design of iterative feature pyramid fusion network

It is well known that feature context information in low levels is are more spatially-sensitive for the features of edges, texture, and corners to be significantly helpful for locating objects but conveys less semantic information. On the contrary, high-level features with low resolution have a large receptive field and are rich context information, being insensitive to geometric distortion and small shift, which is beneficial for dealing with the classification task[34]. Considering that the background in a cluttered environment has the wide coverage of the input image and the little area in a view is occupied by objects, object detection can be viewed as small object detection tasks in cluttered environments. Moreover, object detections suffer from the impact of different sizes and arbitrary orientations. The challenges, as stated above, motivates us to design an iterative feature pyramid network inspired by [35], which integrates three output layers via a down-top and lateral connection because of the smooth feature propagation and multi-time feature reuse. Thus, the proposed IFPF is designed to effectively enhance feature propagation, and fuse multilevel feature reuse is robust against the impact of different sizes (see Fig.4).

The fundamental architecture YOLOv3-Darknet53[6], which considers object detection as a classification and regression problem, can run in real-time and thus, is chosen to be applied to detecting objects rapidly. We employ feature maps $\{Y_1, Y_2, Y_3\}$ generated by the Darknet-53 feedforward network. Note that these three feature maps have strides of 8, 16, 32 pixels with a stride ratio 2. In other words, the stride is equal to the reduction factor of the feature map relative to the original image. In terms of the designed down-top network, we achieve higher resolution features by lateral connections and iterative connections, $\{FM_1, FM_2, FM_3\}$. Since feature maps in the IFPF net are not of the same resolution, we need to normalize the feature maps before integrating them together. For example, to obtain $FM_1$, we first utilize nearest neighbor up-sampling to process all the preceding feature maps for achieving the same size and then, merged it with the next feature maps by concatenation. Also, the merged map and $Y_1$ both undergo a $1 \times 1$ convolutional layer to reduce channel dimensions before element-wise addition. Note that a $1 \times 1$ convolutional filter can be used to change the dimensionality in the filter space. If the number of $1 \times 1$ convolutional channels is bigger than the number of the operated feature map channels then we are increasing dimensionality while if the number of $1 \times 1$ convolutional channels is smaller than the number of the operated feature map channels we are decreasing dimensionality in the filter dimension. Finally, to eliminate the aliasing effects of up-sampling, we append a $3 \times 3$ convolutional layer to the merged map, while reducing the number of channels. After the iteration above, the feature maps are $\{FM_1, FM_2, FM_3\}$. The feature maps $FM_2, FM_3$ are reused two times. During this stage, we still output three feature maps rather than one feature map sharing classification and regression to generate more complementary semantic information.

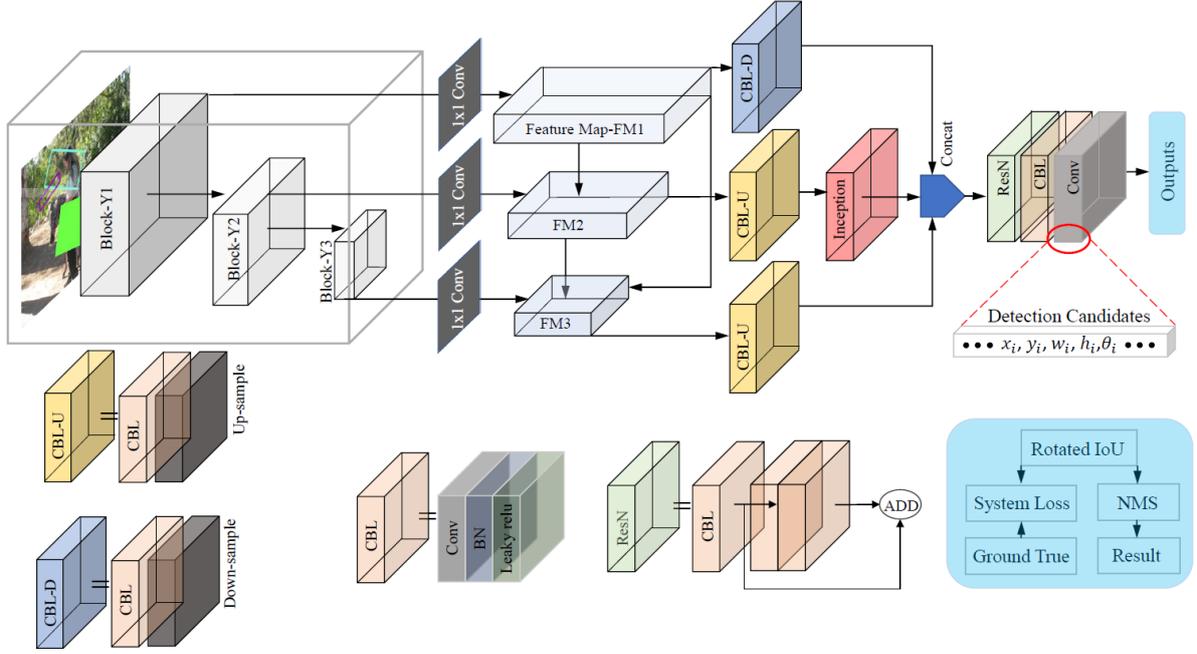

Figure. 3. The overview of the proposed detection model. The backbone is enclosed by the big gray frame. Conv, BN and FM represent the convolutional layer, batch normalization layer and feature map, respectively; IoU and NMS denote the intersection over union and non-maximum suppression, respectively; Concat is concatenation; Inception represents the designed inception net.

*b) Design of the local inception net*

For the purpose of reducing network parameters, we only integrate the local inception model with the middle feature map of the output of the IFPF-Net for the fusion to balance the semantic information and location information while ignoring other less relevant features. The outputs $FM_2$ and $FM_3$ of IFPF-Net are first taken up-sampling to have the same size by the corresponding anchor strides. Then, we pass through the designed inception structure to expand its receptive field and increase semantic information, as shown in Fig.5.

Indeed, the inception structure does not increase the net width but also provides the adaption of the net to the scale. At the same layer, the outputs of different channels are highly correlated. In the inception structure, a $1 \times 1$ convolutional kernel cannot only naturally incorporate features from different channels, but also reduce the number of channels of the input features, thereby reducing the number of parameters included in the module. While other kernels can ensure that the features extracted by them have the scale diversity of the target objects. $3 \times 3$ convolutional kernel, $5 \times 5$ convolutional kernel, and $3 \times 3$ max-pooling kernel are adopted for different convolution branches to achieve feature information extraction for three different scales respectively. Finally, a new feature map $I_2$ is obtained by element-wise addition of the two channels for the middle feature map $FM_2$.

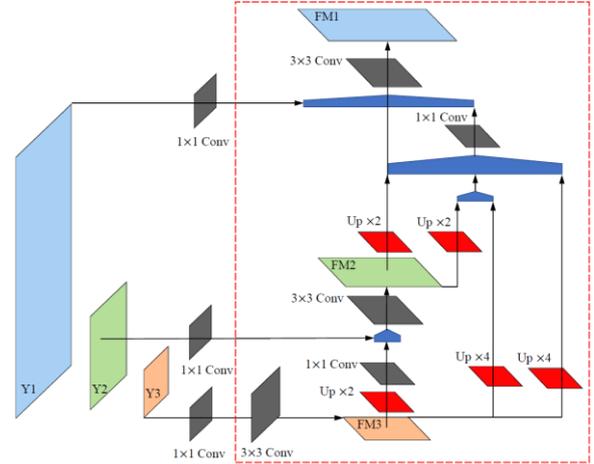

Figure. 4. Design of iterative feature pyramid fusion network. Each feature map is densely connected, and they are merged by the concatenation. The blue five-side polygon represents the concatenation operation.

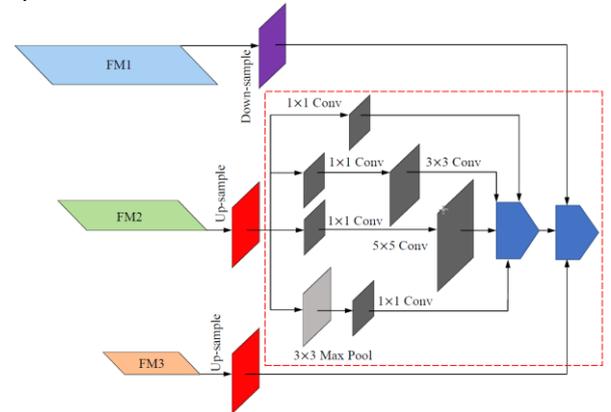

Figure. 5. Design of the local inception net enclosed by the red dashed frame.

*c)* *Design of the multi-receptive-filed feature fusion net*

The multibranch convolution structure above allows object features at different scales to be activated on different branches. Here we integrate feature maps from three branches by adopting reorganization and route layers, which together bring features from the earlier $FM_1$ feature map, the earlier $I_2$ feature map, and the earlier $FM_3$ feature map. Fig. 3 and Fig.5 briefly demonstrate the implementation detail.

We set the size of the integrated feature maps as the same size. In our implementation, we achieve this by the reshape operation which converts larger/smaller feature maps to a set of smaller/larger ones. We set two convolutional layers after the integrated feature map goes through ResN block and CBL block, as shown in Fig.3. Convolutional kernel sizes of these two layers are $3 \times 3$ and $1 \times 1$, respectively. These two convolution layers are the batch normalization layer and rectified linear unit layer sequentially. The final output works as a tensor to predict the location parameters, confidence score, the class score of each rotated bounding box.

*2) Design of prior anchors*

In terms of object detection methods such as YOLO [6], SSD[36], and Faster R-CNN[37], anchors are initial shapes of regions of interest that provide the coarse initial bounding box of the object at each cell in the feature map. Similarly, an ellipse or a rotated rectangle can be estimated coarsely by a prior rotatable anchor. In this case, we add an orientation variable for getting the prior rotated anchor. However, predicting all the five parameters becomes more complex compared with using typical horizontal bounding boxes, not only due to the increased number of unknown variables, but also the fact that the ellipse or rectangle orientation can be arbitrary. But actually, it is fortunate that the shape of an ellipse or rectangle with arbitrary can be assumed to be rotationally invariant, which indicates that it is possible to predict the center, width/major axis, height/minor axis, and orientation independently. To enable a feasible regression, we predict the orientation angle offset relative to prior orientations. We also utilize the k-means algorithm to regress the dimension of ellipses and rectangles in a dataset for obtaining the widths and heights of anchors, as shown in Fig.6. Thus, a series of anchors are calculated by combining different widths, heights, and orientation angles(see Fig.7).

*3) Geometric description of ellipses and rectangles*

To determine an ellipse or rectangle, we need to define the ellipse and rectangle with an arbitrary orientation by five parameters, the geometric center $(x, y)$, the width/major axis, and height/minor axis $(w, h)$ and the orientation angle $\theta$.

Directly predicting parameters results in the model instability[38]. Most of the instability comes from predicting the locations for the box. Similar but different from [38], we do not only predict the 5 parameters $(t_x, t_y, t_w, t_h, t_c)$ of horizontal bounding box and the object class but also regress the orientation angle $t_a$ for obtaining the locations of ellipses and rectangles at each cell in the feature map for the proposed network. As shown in Fig.8, the coordinate $(t_x, t_y)$ is parameterized as an offset, from the top left corner of the feature map by $(c_x, c_y)$, of a particular grid cell and it is normalized between 0 and 1 by the logistic activation. Note that we use an exponential function to constrain the network's predictions of the bounding box width and height to be positive and the anchor box has width and height $(p_w, p_h)$, then the predictions of the bounding box are as follows:

$$\begin{aligned} b_x &= \sigma(t_x) + c_x \\ b_y &= \sigma(t_y) + c_y \\ b_w &= p_w e^{t_w} \\ b_h &= p_h e^{t_h} \end{aligned} \quad (1)$$

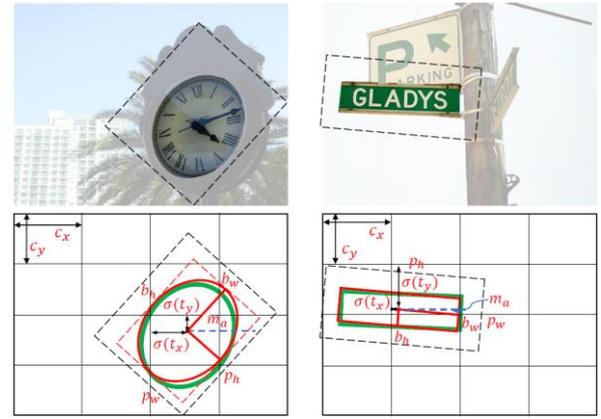

Figure.8. The first row includes the original images with ellipse and rectangle. The green ellipse and rectangle represent the ground truths; the red ellipse and rectangle denote the prediction results.

The orientation angle $\theta$, which is defined as the angle between the semi-major axis direction or the width direction and the x-axis positive direction, determines the orientation of an ellipse or a rectangle. We set the anticlockwise direction to be positive.

The orientation angle needs to be normalized to a limited range so that the prediction for the angle $\theta$ is convergent. Due to the axis symmetry, we need to constrain the angle $\theta$ between $-\frac{\pi}{2}$ and $\frac{\pi}{2}$ to cover all the rotation cases. Although theoretically it may be possible to predict the orientation angle through direct regression, we found that it is impossible to achieve the expected performance. To address this issue, we implement an arc-tangent function for constraining the orientation angle to be between $-\frac{\pi}{2}$ and $\frac{\pi}{2}$ and thus, the potential angle $m_a$ is provided as

$$m_a = \tan^{-1} t_a. \quad (2)$$

The confidence score of each rotated bounding box indicates the probability that the predicted bounding box is an ellipse or rectangle. A logistic activation is applied to normalizing the network's predictions of the confidence score $t_c$ such as to fall in the range between 0 and 1. The confidence score prediction is given as

$$\Pr(class) * IoU_{pre}^{gt} = \sigma(t_c) \quad (3)$$

where $\Pr(class)$ denotes the class probabilities per grid cell, $IoU_{pre}^{gt}$ represents the IoU between the predicted box and the ground truth.

*4) Multi-task loss construction*

Since the proposed network is an end-to-end learning model, we just need to summarize a global loss function for ellipse or rectangle detection. The overall objective loss function is a weighted sum of the rotated bounding box parameters loss and the class confidence loss.

Unlike YOLO[6] and Faster R-CNN[37], we optimize the model global loss $L_{total}$ while training as follows,

$$L_{total} = \lambda_{reg-rbb} L_{reg-rbb}(x_j, \hat{x}_j, y_j, \hat{y}_j, w_j, \widehat{w}_j, h_j, \hat{h}_j) + \lambda_{angle} L_{angle}(\theta_j, \hat{\theta}_j)$$
$$+ \lambda_{obj} L_{obj}(c_j, \hat{c}_j)$$
$$+ \lambda_{noobj} L_{noobj}(c_j, \hat{c}_j)$$
(4)

$$L_{reg-rbb}(x_j, \hat{x}_j, y_j, \hat{y}_j, w_j, \widehat{w}_j, h_j, \hat{h}_j)$$
$$= \sum_{i=0}^{S^2} \sum_{j=0}^{B} \mathbb{1}_{ij}^{obj} [f(x_j, \hat{x}_j) + f(y_j, \hat{y}_j)]$$
$$+ \sum_{i=0}^{S^2} \sum_{j=0}^{B} \mathbb{1}_{ij}^{obj} \left[ \left(\sqrt{w_j} - \sqrt{\widehat{w}_j}\right)^2 + \left(\sqrt{h_j} - \sqrt{\hat{h}_j}\right)^2 \right]$$
(5)

$$L_{angle}(\theta_j, \hat{\theta}_j) = \sum_{i=0}^{S^2} \sum_{j=0}^{B} \mathbb{1}_{ij}^{obj} f(\theta_j, \hat{\theta}_j)$$

$$L_{obj}(c_j, \hat{c}_j) = \sum_{i=0}^{S^2} \sum_{j=0}^{B} \mathbb{1}_{ij}^{obj} f(c_j, \hat{c}_j)$$

$$L_{noobj}(c_j, \hat{c}_j) = \sum_{i=0}^{S^2} \sum_{j=0}^{B} \mathbb{1}_{ij}^{noobj} f(c_j, \hat{c}_j)$$
(6)

where $\mathbb{1}_{ij}^{obj}$ represents that the $j$th bounding box in the cell $i$ is responsible for predicting the target objects and $\mathbb{1}_{ij}^{noobj}$ predicts the case without the target objects; $S$ indicates the number of grid cells divided from the input image; $B$ denotes the rotated bounding box (anchors)number of each grid cell; $\lambda_{reg-rbb}$, $\lambda_{angle}$, $\lambda_{obj}$ and $\lambda_{noobj}$ are the weight parameters of the corresponding loss function $L_{reg-rbb}$, $L_{angle}$, $L_{obj}$, $L_{noobj}$ regarding the rotated bounding box, orientation angle, the cases with and without target objects, respectively. We use these control parameters to adjust the corresponding loss weight; $x_j, y_j, w_j, h_j, \theta_j$ are the location parameters of the prediction bounding box and the corresponding label information is represented by $\hat{x}_j, \hat{y}_j, \widehat{w}_j, \hat{h}_j, \hat{\theta}_j$; $c_j$ and $\hat{c}_j$ represents the prediction parameters and ground truth label information, respectively. For the loss function, we introduce the cross-entropy loss to calculate the losses of the center coordinates, the orientation angle, and the class confidence probabilities.

$$f(X, \hat{X}) = -(1 - \hat{X})\log(1 - \hat{X}) - X\log X \quad (7)$$

where $X, \hat{X}$ represents the prediction result and ground truth,

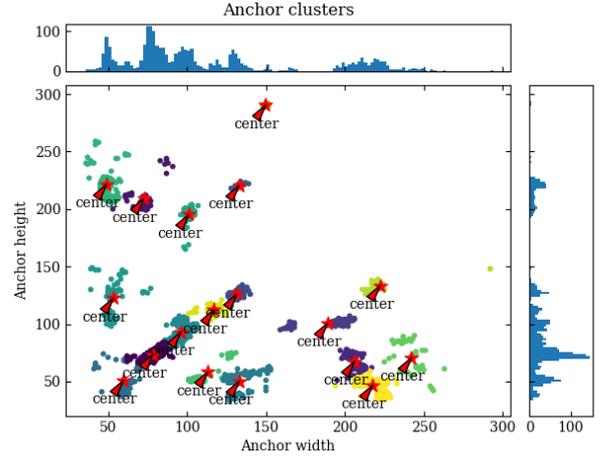

Figure.6. The clusters of prior anchors. The red star represents the centre of each cluster. The zones in different colours denote clusters. The top and right histograms in blue are the frequency histograms of anchor sizes.

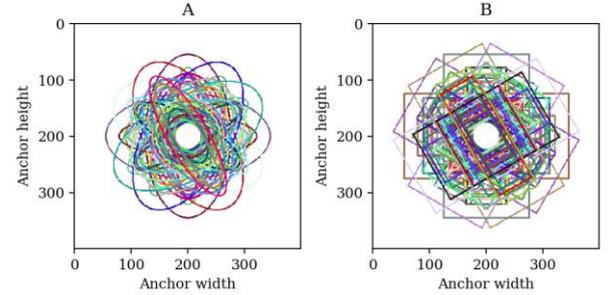

Figure.7. The prior rotated anchors with different sizes (A for ellipses and B for rectangles).

respectively.

*C. Implementation Details*

PyTorch is implemented to train the proposed learning model. We use a step strategy with mini-batch stochastic gradient descent (SGD) to train the networks with the batch size 8 on a GeForce GTX 1070 GPU. A dynamic learning rate is applied to the training process with an initial learning rate of $10^{-3}$ for $\frac{2}{3}$ of total iterations and train for another $\frac{1}{3}$ of total iterations with a decreased learning rate of $10^{-4}$.

## IV. ROBOTIC MANIPULATION EXPERIMENTS

To evaluate the performance of our framework, we run an extensive series of human-robot collaborative manipulation experiments of cylindrical and cubic objects using a stereo camera-ZED, an industrial robot arm-UR5 with the repeatability accuracy of ±0.1 mm, the supported payload of 5kg, the maximum reach of 850mm, and a 3-finger Robotiq gripper without tactile sensor mounted on the robot arm(see Fig.1).

We choose 12 household objects (including cylindrical and cubic objects), covering some appearances. The S-RG dataset, as mentioned above, is used in training the proposed learning model; however, this dataset also includes labeled images. In real scenarios, the tops of cubic objects are not strictly rectangles due to the two cases. One is that the cubic objects are placed at random positions so that the camera observes the cubic objects in arbitrary view angles. The other is that the top surface of objects undertakes distortion to a certain extent due to the squeezing. Contrastively, in synthetic images, all the top shapes of cylindrical and cubic objects are standard ellipses and rectangles. To bridge this gap, we also label 760 images with real scenarios. This synthetic dataset fed to the learning model for robot grasp consists of labeled 760 images and 240 images generated by the proposed solution of generating a synthetic image.

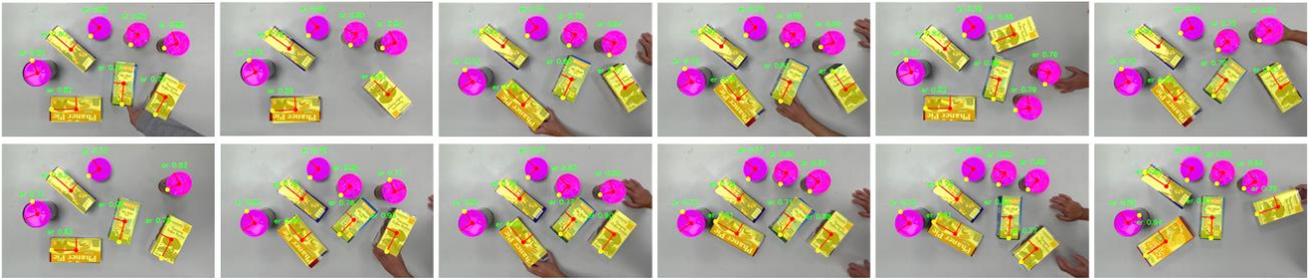

Figure. 9. Some snapshots of a video stream capturing practical scenarios. Each image includes multiple cylindrical and cubic objects with elliptic and rectangle primitives. When the objects are being moved, the proposed detector is still able to detect them, which verifies our detector can be implemented to practical applications.

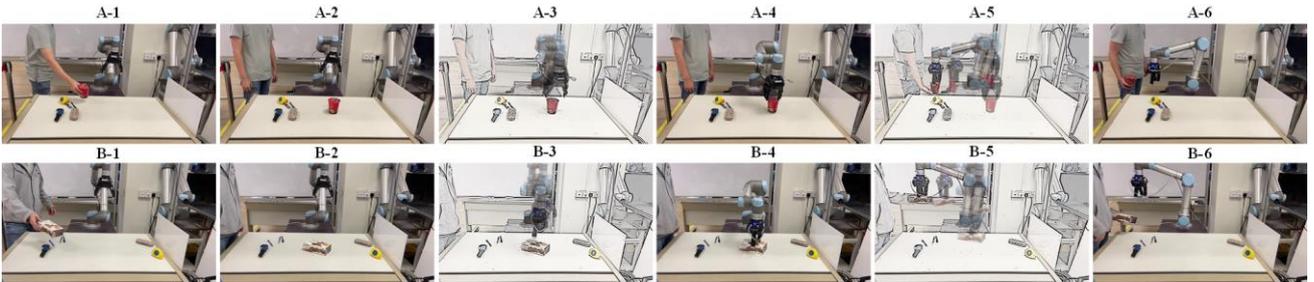

Figure. 10. The workflow of human-robot collaborative manipulation. Feeding an object to the robot (A-1,2 and B-1,2); the grasp motion planning with the trajectory visualization(A-3, B-3); grasping the object(A-4, B-4); the delivering motion planning with the trajectory visualization(A-5, B-5); recovering the object from the robot(A-6, B-6).

*A. Robotic Grasp Synthesis*

Given an image including target objects, our objective is to identify the grasp configuration for robotic manipulation. Since the space information is not involved in the proposed model, we need to introduce a 3D pose representation for the grasp configuration. From an RGB image, the proposed method can detect target objects and obtain ellipse and rectangle centres considered as the 2D pixel coordinates. Some detection cases are indicated in Fig.9. Through registered the depth image, the 2D pixel coordinates of the centres of detected objects are converted to 3D world coordinates. We collect 10-time detection results for each cylindrical or cubic object from the live video stream to get the median value of the centres of 10 ellipses or rectangles, avoiding the cases of partially missing data captured by the camera.

*B. Robotic Manipulations*

*1)Robot grasping a single cylindrical or cubic object*

The first human-robot collaboration task is considered as follows. The vision system first perceives the arrival of new objects via detecting the tops of the cylindrical and cubic objects as ellipses and rectangles on the table and then, the robot subsequently grasps and delivers them to the human operator. While manipulating, the robot is not only a receiver but also a giver. The workflow of collaborative grasp is illustrated in Fig.10. In particular, the 3D centre position of the top of the cylindrical or cubic object is calculated, as introduced above. Inferring the planning paths, the robot performs the actions of grasping the targeted object and passes

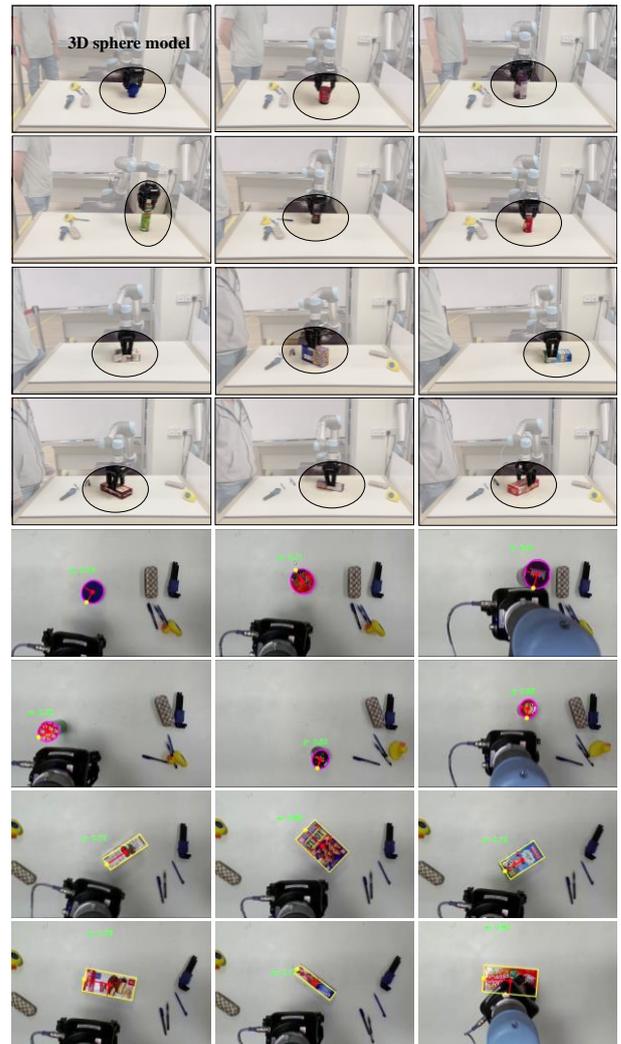

Figure.11. The robotic grasping 6 cylindrical objects and 6 cubic objects highlighted by black ellipse frames captured by the front camera(the first four rows) and by the top ZED camera(the last four rows).

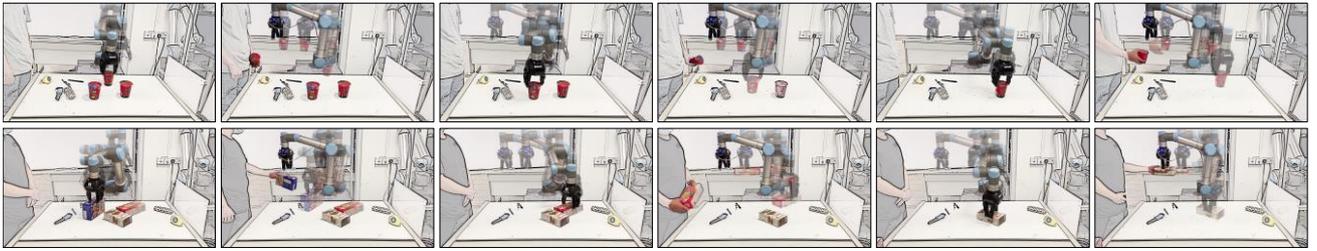

Figure. 12. The robot grasping multiple cylindrical objects(A) and cubic objects(B). The first, second and third trials are illustrated in A-1 and B-1, A-2 and B-2, A-3 and B-3, respectively.

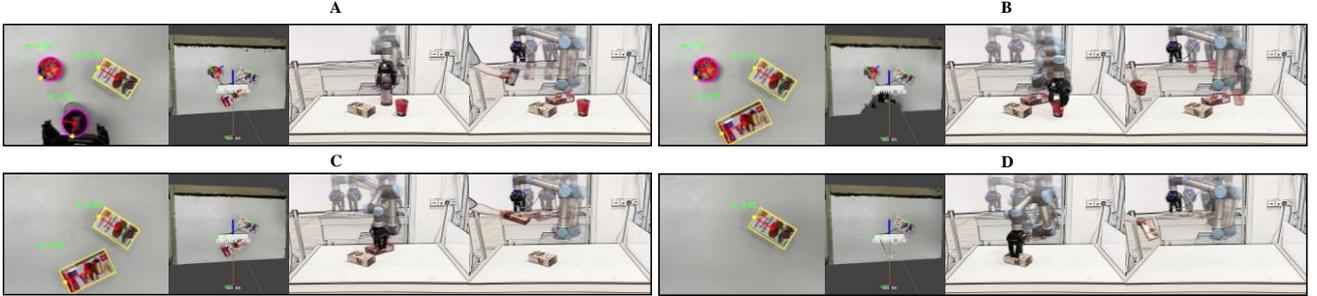

Figure. 13. The robot manipulating multiple mixed cylindrical and cubic objects in the same scenario. Four objects with elliptic and rectangle primitives are placed on the table and the robot runs four trials to complete the grasping round (A, B, C, D). The left figures in A, B, C and D show the grasping perception and pose information in Rviz(ROS) for each trial.

it to the human operator.

We perform at least 120 times of picking experiments for 12 different objects, 6 cylindrical objects including the sphere object, Shincup, chocolate bucket, potato chip bucket, pepper cup, small potato chip cup, and 6 cubic objects such as the coffee box, biscuit box, milk box, pie box, OREO box, swiss roll box, as shown in Fig.11. Each object is put on the table at least ten rounds with random initial orientations and positions. That is, the robot runs at least 10 attempted grasps for per object. When the object is placed close to the left and right top corners of the table, the robot cannot plan a reachable path to achieve a successful grasp due to the large distance between the robot base and the targeted object. If the human operator puts objects within the robotic graspable area, the robot can grasp objects perfectly, which illustrates that the proposed detection strategy can realize accurate ellipse and rectangle detections. Moreover, the appearances of some real objects in robotic grasp experiments are a little or totally different from the scanned 3D models due to product updates. But, fortunately, our self-supervised learning system can track this challenge and successfully adapts to the new scenarios.

*2) Grasps of multiple objects in cluttered environments*

To illustrate that the robot is able to manipulate multiple objects in cluttered environments, we conduct the experiments of the robot grasping cylindrical and cubic objects from a densely packed pile of objects in cluttered environments. The multiple objects are put randomly on a table. The human operator generally feeds objects to the robot by putting them on the table during the period of the robot grasping other objects. Grasping from densely arranged objects presents much more challenging than one single object isolated. Perceiving cluttered visual scenarios and avoiding collisions between the gripper and objects attribute to the main challenges. Our detection approach, however, overcomes the first issue and detects multiple objects at the same time accurately. Providing an accurate perception feedback to the robot is the prerequisite of avoiding collision to realize a successful grasp. We construct a 'closed-loop' grasp system.

In particular, the robot continuously attempts multiple grasps until all objects are grasped, as shown in Fig.12 and Fig.13.

We perform 20 groups of experiments, totally around 100 trials on cylindrical and cubic objects, including 6, 6, 8 groups of grasping the cylindrical objects, cubic objects, mixed cylindrical and cubic objects, respectively. As a result, a total of 18 out of the 20 experimental sets are successfully accomplished without a single failure. Of the two failures, they are due to the estimation of inaccurate positions in the 3D space. Given an accurate detection in the 2D space, the failed grasps are caused by the inaccurate depth information, which allows the gripper to reach an ungraspable position and generates collisions with the object during the grasp attempt. The reflective property of the object's surface results in inaccurate depth. Indeed, the grasp failures cannot attribute to our detector. Under the camera view, the proposed detector almost realizes 100% successful detections. That is, the proposed detector stills manifest a higher average success rate of detecting ellipses and rectangles in cluttered environments, which indicates that the proposed perception strategy can better understand the spatial geometry of the objects.

## V. CONCLUSION

We consider the problem of real-time human-robot collaborative manipulations of cylindrical and cubic objects through the perception strategy in the 2D space. Thus, we present a jointing network that detects ellipses and rectangles with sufficient accuracy and speed to realize robotic manipulation of objects in the presence of complicated scenes. In human-robot collaborative manipulation experiments, the robot successfully grasps and tracks cylindrical and cubic objects in real-time in cluttered environments, which indicates the proposed detector is potentially applied to industrial or household environments. In the future, we will focus on detecting objects with complex shape profiles.